\pdfoutput=1
\documentclass[10pt,twocolumn,letterpaper]{article}

\usepackage[pagenumbers]{wacv} % To force page numbers, e.g. for an arXiv version

\usepackage{graphicx}
\usepackage{amsmath}
\usepackage{amssymb}
\usepackage{booktabs}
\usepackage[utf8]{inputenc} % allow utf-8 input
\usepackage[T1]{fontenc}    % use 8-bit T1 fonts
\usepackage{url}            % simple URL typesetting
\usepackage{booktabs}       % professional-quality tables
\usepackage{amsfonts}       % blackboard math symbols
\usepackage{nicefrac}       % compact symbols for 1/2, etc.
\usepackage{microtype}      % microtypography
\usepackage{times}
\usepackage{soul}
\usepackage{url}
\usepackage[utf8]{inputenc}
\usepackage{multirow}

\usepackage{amsthm}
\usepackage{booktabs}
\usepackage{algorithm}
\usepackage{algorithmic}
\usepackage[switch]{lineno}
\usepackage{amssymb}
\usepackage[normalem]{ulem}
\usepackage{array} 
\usepackage{graphicx}
\usepackage[accsupp]{axessibility}
\usepackage[pagebackref,breaklinks,colorlinks]{hyperref}
\usepackage[normalem]{ulem}
\usepackage{threeparttable}
\usepackage{xurl}
\usepackage{colortbl}
\definecolor{mygray}{gray}{.9}
\definecolor{mypink}{rgb}{.99,.91,.95}
\definecolor{mycyan}{cmyk}{.3,0,0,0}
\definecolor{grayblue}{rgb}{0.7, 0.75, 0.71}
\usepackage{amsmath}    
\usepackage{booktabs}  
\usepackage[pagebackref,breaklinks,colorlinks]{hyperref}

\usepackage[capitalize]{cleveref}
\crefname{section}{Sec.}{Secs.}
\Crefname{section}{Section}{Sections}
\Crefname{table}{Table}{Tables}
\crefname{table}{Tab.}{Tabs.}

\def\wacvPaperID{1729} % *** Enter the WACV Paper ID here
\def\confName{WACV}
\def\confYear{2025}

\begin{document}

\title{Uniform Attention Maps: Boosting Image Fidelity in Reconstruction and Editing}

\author{Wenyi Mo\textsuperscript{1,2}, Tianyu Zhang\textsuperscript{3}, Yalong Bai\textsuperscript{3}, Bing Su\textsuperscript{1,2\thanks{Corresponding Authors.}}, Ji-Rong Wen\textsuperscript{1,2}\\
\textsuperscript{1}Gaoling School of Artificial Intelligence, Renmin University of China \\
\textsuperscript{2}Beijing Key Laboratory of Big Data Management and Analysis Methods \\\textsuperscript{3}Du Xiaoman Technology \\}

\maketitle
% \renewcommand{\thefootnote}{\fnsymbol{footnote}}
% \footnotetext[2]{Corresponding authors.}
\begin{abstract}

Text-guided image generation and editing using diffusion models have achieved remarkable advancements. Among these, tuning-free methods have gained attention for their ability to perform edits without extensive model adjustments, offering simplicity and efficiency. However, existing tuning-free approaches often struggle with balancing fidelity and editing precision. Reconstruction errors in DDIM Inversion are partly attributed to the cross-attention mechanism in U-Net, which introduces misalignments during the inversion and reconstruction process. To address this, we analyze reconstruction from a structural perspective and propose a novel approach that replaces traditional cross-attention with uniform attention maps, significantly enhancing image reconstruction fidelity. Our method effectively minimizes distortions caused by varying text conditions during noise prediction. To complement this improvement, we introduce an adaptive mask-guided editing technique that integrates seamlessly with our reconstruction approach, ensuring consistency and accuracy in editing tasks. Experimental results demonstrate that our approach not only excels in achieving high-fidelity image reconstruction but also performs robustly in real image composition and editing scenarios. This study underscores the potential of uniform attention maps to enhance the fidelity and versatility of diffusion-based image processing methods. Code is available at \href{https://github.com/Mowenyii/Uniform-Attention-Maps}{https://github.com/Mowenyii/Uniform-Attention-Maps}.

\end{abstract}    
\section{Introduction}
\label{sec:intro}

In recent years, the field of image processing has seen significant advancements, particularly with the development of Denoising Diffusion Probabilistic Models (DDPMs)~\cite{ho2020denoising,rombach2022high,saharia2022photorealistic,chen2022re}. These models have revolutionized image composition and editing by enabling more precise and creative control over images~\cite{DBLP:journals/corr/abs-2307-12493,hertz2022prompt}. One of the key innovations has been the introduction of tuning-free methods, which allow for effective editing without the need for extensive model adjustments. These methods offer simplicity and efficiency by manipulating latent vectors during the denoising process, unlocking new possibilities for accurate image editing. However, applying these tuning-free techniques to real-world images presents challenges. In practice, the latent vectors of real images are often unknown, making it difficult to directly apply these methods, which limits their practical use.

\begin{figure}[t]
\centering
\includegraphics[width=1\columnwidth]{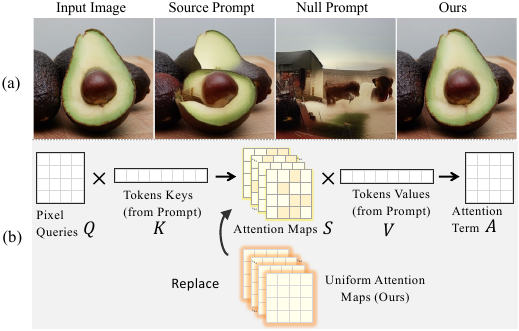} 
\caption{
\textbf{(a)} Image reconstruction using DDIM with different prompts. The first image shows the input image, followed by the reconstruction using the source prompt ``a photo of avocados," the null prompt (an empty string), and the result using Uniform Attention Maps combined with token values from the null prompt. \textbf{(b)} Our approach introduces Uniform Attention Maps, where traditional attention maps are replaced with uniform maps that distribute attention weights equally across the token dimension. By combining these uniform maps with the value tokens \( V \), we generate a more balanced attention term \( A \). This method ensures consistent attention, resulting in more accurate image reconstructions, as demonstrated in the final image of part~(a).
}
\label{fig:motivation}
 \vskip -0.1in
% \vspace{-0.7cm} 
\end{figure}

\begin{table*}[t]
\small
\centering
\resizebox{0.9\linewidth}{!}{
\begin{tabular}{lcccccc}
\toprule

            Method  & \begin{tabular}[c]{@{}l@{}}Base \\ Model\end{tabular}& \begin{tabular}[c]{@{}l@{}}Structure $\downarrow$\\ Distance$_{\times10^{3}}$\end{tabular} & PSNR$\uparrow$ & LPIPS$_{\times10^{3}}$$\downarrow$& MSE$_{\times10^{4}}$$\downarrow$& SSIM$_{\times10^{2}}$$\uparrow$\\\midrule %\hline%

Upper Bound & VQAE~\cite{esser2021taming} & 2.39 & 28.58&34.20 &21.57&82.04\\  \hline

Null Prompt  & SD 1.4 & 15.31                            & 22.88            & 124.35              & 69.60              & 72.18            \\
Source Prompt & SD 1.4 & 11.31                            & 23.89            & {101.47}     & 55.43             & 74.45            \\

Zero Cross-Attention Maps  & SD 1.4 & {11.13}& {24.36} & 102.83 & {51.17}& {74.97}\\
TF-ICON~\cite{DBLP:journals/corr/abs-2307-12493} & SD 1.4  & {5.51} & {25.57  } & {64.12} & {37.34} & {77.70}   \\
\rowcolor[gray]{0.9} Uniform Attention Maps (Null)  & SD 1.4 & \uline{4.76} & \textbf{26.97  } & \uline{57.29} & \textbf{28.98} & \uline{79.29}   \\
\rowcolor[gray]{0.9} Uniform Attention Maps (Src)  & SD 1.4 & \textbf{4.67} & \uline{26.96  } & \textbf{54.17} & \uline{29.05} & \textbf{79.33}   \\

\bottomrule
\end{tabular} }
\vskip -0.08in
\caption{Reconstruction performance on the PIE benchmark~\cite{DBLP:journals/corr/abs-2310-01506} using DDIM Inversion with $20$ timesteps under various conditions without CFG. Our method, Uniform Attention Maps, achieves higher fidelity to the original image than others. Additionally, the reconstruction results using token values from source and null prompts are similar, demonstrating the robustness of our approach across different prompts.
}  \label{tab:novelty}
\vskip -0.1in
\end{table*}

To overcome this, researchers have developed inversion methods like Denoising Diffusion Implicit Models (DDIM) Inversion~\cite{song2021denoising}, which map images back to their noisy latent vectors using a trained diffusion model. This approach has been particularly effective for unconditional diffusion models. Additionally, recent advances in text-conditioned DDIM inversion~\cite{mokady2022null,miyake2023negative,DBLP:journals/corr/abs-2306-05414,DBLP:journals/corr/abs-2310-01506} have further improved image editing by incorporating classifier-free guidance (CFG)~\cite{ho2021classifier} during the generation and editing stages. These enhancements have led to more effective edits, but challenges remain. Current methods still struggle to balance preserving the original image details with making user-defined changes.

Existing methods~\cite{mokady2022null,miyake2023negative,DBLP:journals/corr/abs-2310-01506,cao2023masactrl} typically use a dual-branch approach after inverting input images, separating the process into reconstruction (source) and editing (target) branches. While this approach has yielded impressive results, it also introduces challenges, such as discrepancies between noise predictions in the inversion and reconstruction phases in the reconstruction branch, which can lead to the loss of important image details~\cite{wallace2022edict}. Various strategies have been proposed to address these issues. Some approaches, like Null-text Inversion~\cite{mokady2022null}, use optimization techniques to minimize the distance between the representations between the reconstruction and inversion phases. On the other hand, methods like Proximal Guidance~\cite{DBLP:journals/corr/abs-2306-05414} improve reconstruction effectiveness by introducing  an extra regularization term, without extensive tuning. Despite these advancements, the reconstruction effectiveness varies significantly with different prompts. As shown in Fig.~\ref{fig:motivation}~(a), reconstruction outcomes can differ significantly based on these conditions. This leads us to the core questions of our research: Given the assumption of DDIM inversion with adjacent noise prediction approximation, why do different conditions lead to varied reconstruction outcomes? How can we improve image reconstruction effectiveness in text-conditioned scenarios?

To address these questions, our study focuses on the cross-attention mechanism within the U-Net architecture used in diffusion models. We are the first to analyze DDIM inversion and reconstruction under text-conditioned settings from a structural perspective. Our findings reveal that cross-attention plays a pivotal role in the reconstruction errors observed in current methods. To address this, we propose an improved image reconstruction method that leverages uniform cross-attention to enhance the effectiveness of text-conditioned image reconstruction and composition. Additionally, we introduce an automatic mask generation technique to improve the performance of existing image editing algorithms, making our approach more robust and applicable to a wider range of scenarios.

Our contributions are threefold: (1) We provide a detailed analysis of how cross-attention impacts image reconstruction, (2) We propose an enhanced reconstruction method that shows superior performance in both image composition and editing tasks, and (3) We develop an automatic mask generation technique that significantly improves the accuracy and effectiveness of image editing. Through these innovations, we aim to advance image processing, offering new tools and methods that can be easily adopted in practical applications.

\section{Related work}

\begin{figure*}[t]
	
	\begin{center}
		\centering
		{\includegraphics[width=2\columnwidth]{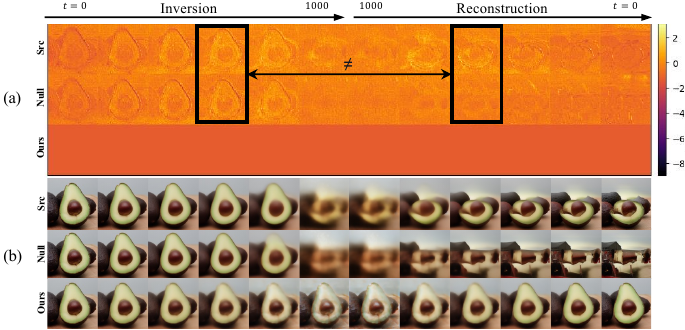}}
	\vskip -0.05in
		\caption{The process of reconstruction using DDIM inversion under various conditions. It visually depicting (a) the heatmaps of the cross-attention term  $A^{(l)}$, summed along the dimension $d^{(l)}_x$, from the U-Net model’s layers with output dimensions of  $64 \times 64$, and (b) the predicted latent representation \( \hat{z}_0 \) at different stages of both the inversion and reconstruction processes. In (a), discrepancies in the cross-attention maps between the inversion and reconstruction phases are evident, with misalignment causing errors in image fidelity under the source and null prompt conditions.   In (b), the reconstructed images show significant distortions under the source and null conditions, whereas our method consistently maintains high image quality throughout the reconstruction process.}
\vskip -0.08in

		\label{fig:visual_ddim_inv_fail_fruit}
	\end{center}
	\vskip -0.2in
\end{figure*}

In recent years, significant advancements have been made in the field of text-guided vision tasks, encompassing areas such as vision-language inference~\cite{Radford2021LearningTV,Li2022SupportingVM,DBLP:conf/mm/0002HZP23,DBLP:conf/aaai/CuiPWZZ24}, text-to-image generation~\cite{saharia2022photorealistic,rombach2022high,mo2024dynamic,sd3}, and image editing~\cite{hertz2022prompt,cao2023masactrl,mokady2022null,miyake2023negative,DBLP:journals/corr/abs-2310-01506}.  While our focus in this paper is on text-conditioned image editing with diffusion-based models, these works highlight the broader importance of effective text guidance in vision-related tasks. The biggest challenge in this task is how to achieve the intention of the guiding texts while ensuring fidelity to the input image. Previous works can be categorized as end-to-end editing models, tuning-based methods, attention-based methods, and sample-based methods. (a) End-to-End Editing Model: Methods like InstructPix2Pix~\cite{brooks2022instructpix2pix} and DiffusionCLIP~\cite{kim2022diffusionclip} fine-tune pre-trained text-to-image models to revise images based on simple instructions, allowing for efficient and quick edits without per-example fine-tuning or inversion. (b) Tuning-based methods: Tuning-based methods involve training a set of learnable parameters or fine-tuning a model to encapsulate certain concepts. Methods such as Imagic~\cite{kawar2022imagic} and Unitune~\cite{Valevski2022UniTuneTI} specifically fine-tune the model on the input image to achieve high fidelity. 
These methods are time-consuming and the misalignment of learned variables with the diffusion model's expected input distribution compromises the integrity and quality of edits, limiting their practical use in fast-processing and high-fidelity applications~\cite{DBLP:journals/corr/abs-2310-01506}.
(c) Attention-based methods: Attention mechanisms allow models to ``focus'' on specific parts of an image, making it possible to edit certain areas or aspects without affecting the entire image. These methods improve precision, context awareness, and efficiency of image editing, enabling more complex edits.
For instance, Prompt-to-Prompt~\cite{hertz2022prompt} and MasaCtrl~\cite{cao2023masactrl} focus on integrating attention mechanisms to ensure that edits are contextually aware and maintain the essence of the input image. 
Our method can be combined with them to help achieve better reconstruction results and enhance editing efficiency. 
(d) Sample-based methods: Methods like Null-text Inversion~\cite{mokady2022null}, Negative-prompt Inversion~\cite{miyake2023negative}, Proximal Guidance~\cite{DBLP:journals/corr/abs-2306-05414}, Direct Inversion~\cite{DBLP:journals/corr/abs-2310-01506}, EDICT~\cite{wallace2022edict}, and Edit Friendly DDPM~\cite{HubermanSpiegelglas2023AnEF} focus on refining the reconstruction process to improve the fidelity of the input image during editing. 
TF-ICON~\cite{DBLP:journals/corr/abs-2307-12493} shows that semantically meaningful text in the input prompt introduces deviations in the diffusion process, causing a mismatch between the forward and reverse trajectories in the ODE-based sampling steps. To address this, the concept of an ``exceptional prompt'' is introduced, using a selected token to stabilize the diffusion process and improve image reconstruction. However, this approach often struggles to generalize across generative models due to inherent differences in their architectures, especially in text encoders. DiffEdit~\cite{Couairon2022DiffEditDS} uses differences in noise predictions to create masks for faithful image editing. We also use masks during editing. The proposed adaptive masks vary with each timestep to better align with our reconstruction method and achieve superior editing performance.

\section{Method}
\label{sec:method}

In this section, we investigate the underlying causes of reconstruction errors associated with different prompts and propose a method to improve reconstruction by reducing the impact of the cross-attention term. We then introduce an automatic mask generation technique that integrates this method into existing image editing algorithms.

\subsection{Preliminaries}
\noindent\textbf{DDIM Inversion.} Denoising Diffusion Implicit Models (DDIMs)~\cite{song2021denoising} are an extension of Denoising Diffusion Probabilistic Models (DDPMs)~\cite{ho2020denoising,rombach2022high,saharia2022photorealistic,chen2022re}, designed to offer a deterministic sampling process. The reverse process in DDIM can be described as follows:
\begin{equation}
\begin{split}
z_{t-1}=& {\sqrt{1-\alpha_{t-1}} \cdot \boldsymbol{\epsilon}_\theta(z_t, t)}+\\&\sqrt{\alpha_{t-1}} \underbrace{\left(\frac{z_{t}-\sqrt{1-\alpha_{t}} \boldsymbol{\epsilon}_\theta(z_t, t)}{\sqrt{\alpha_{t}}}\right)}_{\text {`` predicted } \hat{z}_{0,t} \text { '' }},
\end{split}
\label{eq:ddim}
\end{equation}
where \( z_{t-1} \) represents the latent vector at the previous timestep, derived from \( z_t \) at the current timestep. \(\hat{z}_{0,t}\) denotes the estimated clean image at timestep \( t \). The parameters \( \alpha_t \) are derived from the forward diffusion process, and the function \( \boldsymbol{\epsilon}_\theta(z_t, t) \) estimates the noise at each timestep. To make this process more practical for image editing, we can rearrange Eq.~(\ref{eq:ddim}) as:
\begin{equation}
\begin{split}
z_{t}=&{\sqrt{1-\alpha_{t}} \cdot \boldsymbol{\epsilon}_\theta(z_t, t)}+\\&\sqrt{\alpha_{t}} {\left(\frac{z_{t-1}-\sqrt{1-\alpha_{t-1}} \boldsymbol{\epsilon}_\theta(z_t, t)}{\sqrt{\alpha_{t-1}}}\right)}.
\end{split}
\label{eq:ddim_re}
\end{equation}
When applying this model to real images, the goal is to obtain the initial noise vector \( z_T \) from a given image representation \( z_0 \) as the starting point for further editing. However, directly computing \( z_t \) requires the noise prediction \( \boldsymbol{\epsilon}_\theta(z_t, t) \), which is not always accessible. Therefore, during the inversion process, an approximation is made by using the noise prediction from the previous timestep \( \boldsymbol{\epsilon}_\theta(z_{t-1}, t-1) \)~\cite{wallace2022edict}.
This approach results in a sequence of latent variables, \( \{z^*_t\}_{t=1}^T \), that traces back through the diffusion process:
\begin{equation}
\begin{split}
z^*_{t}=&{\sqrt{1-\alpha_{t}} \cdot \boldsymbol{\epsilon}_\theta(z^*_{t-1}, t-1)}+\\&\sqrt{\alpha_{t}} \underbrace{\left(\frac{z^*_{t-1}-\sqrt{1-\alpha_{t-1}} \boldsymbol{\epsilon}_\theta(z^*_{t-1}, t-1)}{\sqrt{\alpha_{t-1}}}\right)}_{\text {`` predicted } \hat{z}_{0,t} \text { '' }}.
\end{split}
\label{eq:ddim_inversion}
\end{equation}

\noindent\textbf{Cross-attention mechanism.} 
In diffusion models implemented using U-Net, the text condition is typically incorporated through a cross-attention mechanism~\cite{rombach2022high, saharia2022photorealistic}. When predicting \( \boldsymbol{\epsilon}_\theta(z_t, t, \mathbf{c}) \), where \( \mathbf{c} \in \mathbb{R}^{N \times d_c} \) represents the input text and \( N \) is the token number of the input text, the flattened intermediate representation of the \( l^{\text{th}} \) layer of the model \( \boldsymbol{\epsilon}_\theta \) at time step \( t \), denoted as \( x^{(l)}_t \in \mathbb{R}^{M^{(l)} \times d^{(l)}_x} \), is updated via cross-attention as follows:
\begin{equation}
\tilde{x}^{(l)}_t = x^{(l)}_t + A^{(l)}_t,
\label{eq:attention1}
\end{equation}
where $\tilde{x}^{(l)}_t$ is the updated representation, and $A^{(l)}_t$ represents the cross-attention term (or update term), calculated as:
\begin{equation}
A^{(l)}_t = S^{(l)}_t\cdot V^{(l)},
\label{eq:attention2}
\end{equation}
with the score map $S^{(l)}_t \in \mathbb{R}^{M^{(l)} \times N}$ defined by:
\begin{equation}
S^{(l)}_t = \mathrm{softmax}\left(\frac{Q^{(l)}_t(K^{(l)})^T}{\sqrt{d}}\right),
\label{eq:score_maps}
\end{equation}
where $Q^{(l)}_t \in \mathbb{R}^{M^{(l)} \times d^{(l)}}$ is the linear transformation of $x^{(l)}_t$, and $K^{(l)}, V^{(l)} \in \mathbb{R}^{N \times d^{(l)}}$ are the linear transformations of $\mathbf{c}$. Note that $K^{(l)}$ and $V^{(l)}$ are independent of the time step.

\begin{figure}[t]
	
	\begin{center}
		\centering
		{\includegraphics[width=0.9\columnwidth]{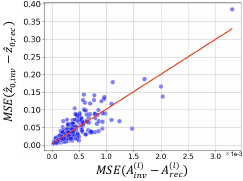}}
		\vskip -0.1in
		\caption{Correlation between MSE of cross-attention term $A^{(l)}$ and clean image prediction $\hat{z}_{0}$ during inversion and reconstruction. The scatter plot shows that discrepancies in the cross-attention term \( A^{(l)}_t \) from all U-Net model's layers with output dimensions of $64\times64$ during the inversion and reconstruction phases contribute significantly to the Mean Squared Error (MSE) in the predicted clean image \( \hat{z}_{0,t} \), as evidenced by the positive correlation across 700 images from the PIE benchmark~\cite{DBLP:journals/corr/abs-2310-01506}.}
		\label{fig:point}
	\end{center}
\vskip -0.21 in
\end{figure}

\subsection{The Devil in Reconstruction: Non-uniform Cross-attention}

DDIM inversion assumes that the noise predictions at adjacent timesteps,  $\boldsymbol{\epsilon}_{\theta}(z_t, t)$  and $\boldsymbol{\epsilon}_{\theta}(z_{t-1}, t-1)$, are approximately equal. When conditioned on a prompt  $\mathbf{c}$,  the difference between \(\boldsymbol{\epsilon}_{\theta}(z_t, t, \mathbf{c})\) and \(\boldsymbol{\epsilon}_{\theta}(z_{t-1}, t-1, \mathbf{c})\) becomes significant, leading to notable reconstruction errors. These discrepancies arise because the cross-attention term  $A^{(l)}$ , which integrates semantic guidance from the prompt into the intermediate latent representation, is misaligned between the inversion and reconstruction processes. 

To quantify this phenomenon, we analyze 700 images from the PIE benchmark to explore the relationship between the Mean Squared Error (MSE) of the predicted clean image \( \hat{z}_{0,t} \) and the cross-attention term \( A^{(l)}_t \) during the inversion and reconstruction phases. Detailed experimental settings can be found in the supplementary materials. As illustrated in Fig.~\ref{fig:point}, the scatter plot highlights this relationship, with a clear positive correlation shown by the red trend line. This indicates that discrepancies in the cross-attention term \( A^{(l)}_t \)  contribute to errors in the reconstructed image \( \hat{z}_{0,t} \).

This observation is further supported by the visualization experiments presented in Fig.~\ref{fig:visual_ddim_inv_fail_fruit}, which track the inversion and reconstruction trajectories for an avocado example. At each timestep, we first compute the update term \( A^{(l)} \) from the U-Net model’s layers, as illustrated in Fig.~\ref{fig:visual_ddim_inv_fail_fruit}~(a). Following this, the clean predicted image \( \hat{z}_{0,t} \) is generated, as shown in Fig.~\ref{fig:visual_ddim_inv_fail_fruit}~(b). Fig.~\ref{fig:visual_ddim_inv_fail_fruit}~(a) highlights a clear mismatch between inversion and reconstruction, particularly under source and null prompt conditions (black-boxed regions), suggesting that misalignment in the cross-attention mechanism contributes to these distortions. The observed misalignment of the update term \( A^{(l)} \) across both trajectories at the same timestep in Fig.~\ref{fig:visual_ddim_inv_fail_fruit} suggests that cross-attention is responsible for the reconstruction errors. Experiment details can be found in the appendix.

\subsection{Our solution}
\subsubsection{Uniform Cross-attention Maps}

Experimentally, the interaction between text prompts and the model’s intermediate representation using the attention mechanism introduces inconsistencies that degrade the quality of the final image reconstruction.

Building on our experiments and analyses, we propose Uniform Cross-attention Maps to enhance stability and consistency across various prompts and models. Instead of relying on traditional cross-attention maps, which vary significantly depending on the input prompt, we introduce uniform attention maps where each element is assigned a fixed value of \( 1/N \):
\begin{equation}
S^{(l)}_{uniform} =  \frac{1}{N} \mathbf{1}_{ M^{(l)} \times N},
\end{equation}
Here, \( \mathbf{1}_{M^{(l)} \times N} \) denotes an \(M^{(l)} \times N\) matrix with all elements equal to $1$, with $M^{(l)}$ being the number of visual tokens and $N$ the number of conditioning tokens.
This uniform distribution of attention reduces the variance introduced by different text prompts, ensuring that the model's focus remains balanced across all tokens in \(x^{(l)}\). As demonstrated in Fig.~\ref{fig:motivation}~(a), our approach effectively mitigates the deviations caused by semantic variations in text prompts, resulting in more reliable and consistent image reconstructions, as evidenced by the improved performance metrics in~\cref{tab:novelty,tab:recon_celeb}. In contrast, Zero Cross-Attention Maps, which replace the cross-attention term \( A^{(l)} \) with zeros, eliminate all semantic guidance from text prompts. While this ensures consistency, it leads to overly simplistic reconstructions and disrupts the pretraining distribution of latent features \( x^{(l)} \), which were optimized to interact with cross-attention. This deviation significantly degrades the model's ability to preserve fine-grained details and complex structures. These limitations underscore the importance of uniform attention maps, which not only reduce prompt variance but also maintain compatibility with the pretraining distribution to achieve high-fidelity reconstructions.

\begin{figure}[t]
	
	\begin{center}
		\centering
		{\includegraphics[width=1\columnwidth]{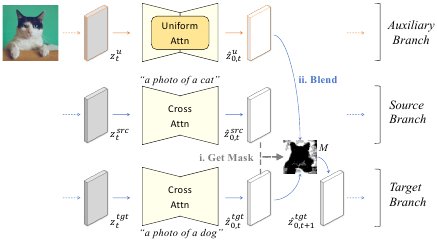}}
		\vskip -0.1in
		\caption{The proposed tuning-free image editing framework. We find that using Uniform Cross-attention Maps yields excellent reconstruction results, as shown in Tab.~\ref{tab:novelty}. We introduce an auxiliary branch and generate masks based on the differences between the source branch and the target branch to blend the results of the auxiliary branch. Our method effectively enhances the performance of existing image editing algorithms. The process of using Uniform Attention Maps is shown in~\cref{fig:motivation}~(b).}
		\label{fig:method_pipe}
	\end{center}
	\vskip -0.2in
\end{figure}

\subsubsection{Adaptive Mask Guided Editing}

The direct use of uniform attention maps in current text-driven editing pipelines presents challenges, as these pipelines typically rely on manipulating cross-attention maps to achieve precise edits. However, the exceptional reconstruction performance of uniform attention maps offers a unique opportunity to improve editing tasks. To harness this reconstructive capability, we propose a novel approach, namely adaptive mask-guided editing, which effectively utilizes the strengths of uniform attention maps in editing scenarios. The overall process is illustrated in Fig.~\ref{fig:method_pipe}.

In this method, the input image is processed through three parallel branches: the auxiliary branch, the source branch, and the target branch. The auxiliary branch, which uses a null prompt combined with uniform cross-attention maps, ensures stable reconstruction. The source branch uses the source prompt \( \mathbf{c}_{src} \), while the target branch operates with the target prompt \( \mathbf{c}_{tgt} \) to apply the desired edits.

To further refine this process, we introduce an adaptive mask generation technique that compares the noise predictions between the source and target branches. This comparison yields a difference, \( \mathit{diff}_t = \left| \hat{z}_{0,t}^{tgt} - \hat{z}_{0,t}^{src} \right| \), identifying areas requiring modification. A threshold \( \lambda \) is then applied to this difference to create a mask \( M \), which is subsequently refined using a dilation operation with a square kernel to handle minor inconsistencies:
\[
M = dilate(|\hat{z}^{tgt}_{0,t} - \hat{z}^{src}_{0,t}| \leq \lambda).
\]
After \( T_{mask} \) timesteps, this mask is employed to blend the predicted clean images $\hat{z}^{u}_{0,t}$ and $\hat{z}^{tgt}_{0,t}$ from the auxiliary and target branches, ensuring that the model preserves key details from the original image while applying targeted edits:
\[
\hat{z}^{tgt}_{0,t} = M \odot \hat{z}^{u}_{0,t} + (1-M) \odot \hat{z}^{tgt}_{0,t}.
\]

By selectively blending the clean images using the mask, the algorithm achieves a balance between maintaining the original image’s fidelity and incorporating the desired modifications. This approach ensures that critical details are preserved, while the edits are seamlessly integrated into the final output. For a detailed representation of the algorithm, please refer to the pseudo-code in supplementary materials.

\begin{table}[t]
	\small
	\begin{center}
		\resizebox{0.48\textwidth}{!}{
			\begin{tabular}{llccc}
				\toprule
				& Method & MAE $\downarrow$ & LPIPS $\downarrow$ & SSIM $\uparrow$ \\ 
				\midrule
    				Upper Bound & VQAE \cite{esser2021taming} & 0.018 & 0.043 & 0.919 \\ \midrule
				\multirow{7}{*}{Diffusion} & SD w/ CFG & 0.134 & 0.340 & 0.637  \\
				& SD w/ Cond. & 0.126 & 0.308 & 0.654 \\
				& SD w/ Uncond. & 0.126 & 0.304 & 0.655 \\

				& TF-ICON~\cite{DBLP:journals/corr/abs-2307-12493} & {0.019} & {0.047} & {0.918}  \\ \cmidrule{2-5}
    				& TF-ICON*~\cite{DBLP:journals/corr/abs-2307-12493} &0.021 & 0.045&0.834  \\
            				& UAM* & \textbf{0.019} &\textbf{0.041} &\textbf{0.839} \\
                            \bottomrule
			\end{tabular}
            }
	\end{center}
 \vskip -0.2in
 	\caption{Quantitative comparison of image reconstruction on CelebA-HQ~\cite{karras2017progressive}. Additional experimental results and setting details are  in~\cite{DBLP:journals/corr/abs-2307-12493}. Methods marked with `*' indicate results on A800.}
	\label{tab:recon_celeb}
     \vskip -0.1in
\end{table}

\begin{figure*}[t]
\centering
\includegraphics[width=2\columnwidth]{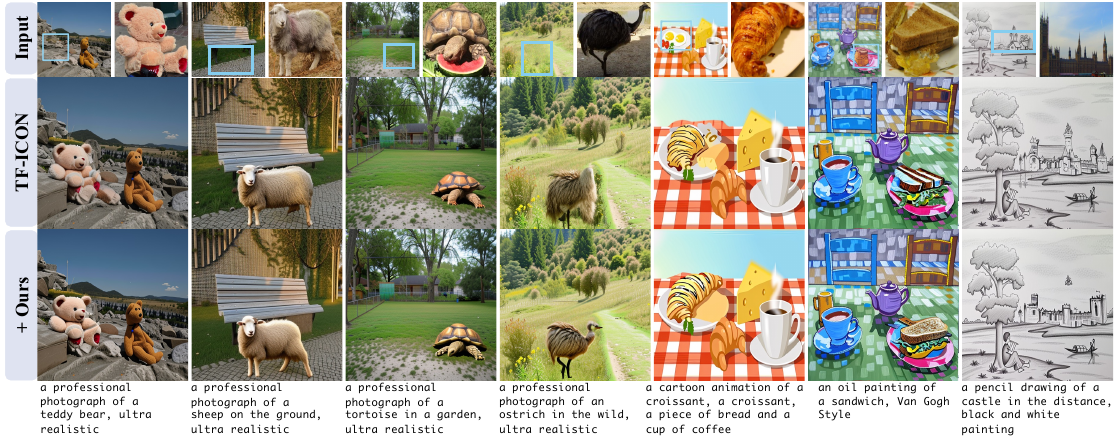} 
\vskip -0.1in
\caption{Qualitative comparison with SOTA and baselines in image composition task on TF-ICON bench mark. Our method generates images with higher fidelity to the reference images and produces more realistic results.}
\label{fig:cmp_tficon}
 \vskip -0.15in
\end{figure*}

\begin{table*}[t]
\small
    \centering

\begin{tabular}{c|c|cccc|cc}
\toprule
     \multirow{2}{*}{Method }      & Structure $\downarrow$ & \multicolumn{4}{c}{Background Preservation} \vrule& \multicolumn{2}{c}{CLIP Score} \\\cline{3-6}\cline{7-8} 
     & Distance {$\times10^{3}$} & PSNR$\uparrow$ & LPIPS {$\times10^{3}$}$\downarrow$& MSE {$\times10^{4}$}$\downarrow$& SSIM {$\times10^{2}$}$\uparrow$& Whole$\uparrow$  & Edited$\uparrow$ \\\midrule
     
  DDIM & 28.38 & 22.17 & 106.62 & 86.97 & 79.67 & 23.96 & 21.16 \\
\rowcolor[gray]{0.9} + Ours & {24.80}$_{\textcolor{red}{13\%\downarrow}}$ & \textbf{22.96}$_{\textcolor{red}{3.6\%\uparrow}}$ & {91.56}$_{\textcolor{red}{14.1\%\downarrow}}$ & \textbf{76.17}$_{\textcolor{red}{12.4\%\downarrow}}$ & {81.19}$_{\textcolor{red}{1.9\%\uparrow}}$ & {24.29}$_{\textcolor{red}{1.4\%\uparrow}}$ & {21.21}$_{\textcolor{red}{0.2\%\uparrow}}$ \\

\hline\hline
DI & 24.70 & 22.64 & 87.94 & 81.09 & 81.33 & 24.38 & 21.35 \\

\rowcolor[gray]{0.9} + Ours & \textbf{24.60}$_{\textcolor{red}{0.4\%\downarrow}}$ & {22.68}$_{\textcolor{red}{0.2\%\uparrow}}$ & \textbf{87.39}$_{\textcolor{red}{0.6\%\downarrow}}$ & {80.63}$_{\textcolor{red}{0.6\%\downarrow}}$ & \textbf{81.52}$_{\textcolor{red}{0.2\%\uparrow}}$ & \textbf{24.59}$_{\textcolor{red}{0.9\%\uparrow}}$ & \textbf{21.46}$_{\textcolor{red}{0.5\%\uparrow}}$ \\

  \hline\hline
\end{tabular}
\vskip -0.08in
\caption{Quantitative comparison of image editing on the PIE benchmark. The methods are compared using the Masactrl attention control~\cite{cao2023masactrl}. }  \label{tab:masa}

\end{table*}

\begin{table*}[t]
\small
    \centering

\begin{tabular}{c|c|cccc|cc}
\toprule
     \multirow{2}{*}{Method }      & Structure $\downarrow$ & \multicolumn{4}{c}{Background Preservation} \vrule& \multicolumn{2}{c}{CLIP Score} \\\cline{3-6}\cline{7-8} 
     & Distance {$\times10^{3}$} & PSNR$\uparrow$ & LPIPS {$\times10^{3}$}$\downarrow$& MSE {$\times10^{4}$}$\downarrow$& SSIM {$\times10^{2}$}$\uparrow$& Whole$\uparrow$  & Edited$\uparrow$ \\\midrule

NT &13.44 &27.03& 60.67 &35.86 &84.11& 24.75 &21.86\\
NP &16.17 &26.21 &69.01 &39.73 &83.40& 24.61 &21.87\\
StyleD  &11.65 &26.05 &66.10& 38.63 &83.42& 24.78& 21.72\\ \hline

  DDIM & 69.43 & 17.87 & 208.80 & 219.88 & 71.14 & 25.01 & \textbf{22.44} \\

\rowcolor[gray]{0.9} + Ours & {49.78}$_{\textcolor{red}{28.3\%\downarrow}}$ & {18.97}$_{\textcolor{red}{6.2\%\uparrow}}$ & {180.85}$_{\textcolor{red}{13.4\%\downarrow}}$ & {181.95}$_{\textcolor{red}{17.2\%\downarrow}}$ & {73.33}$_{\textcolor{red}{3.1\%\uparrow}}$ & {25.09}$_{\textcolor{red}{0.3\%\uparrow}}$ & {22.23} \\ \hline
DI & 11.65 & 27.22 & 54.55 & 32.86 & 84.76 & 25.02 & 22.10 \\

\rowcolor[gray]{0.9} + Ours & \textbf{11.05}$_{\textcolor{red}{5.2\%\downarrow}}$ & \textbf{27.44}$_{\textcolor{red}{0.8\%\uparrow}}$ & \textbf{52.17}$_{\textcolor{red}{4.4\%\downarrow}}$ & \textbf{31.46}$_{\textcolor{red}{4.3\%\downarrow}}$ & \textbf{85.15}$_{\textcolor{red}{0.5\%\uparrow}}$ & \textbf{25.17}$_{\textcolor{red}{0.6\%\uparrow}}$ & {22.14}$_{\textcolor{red}{0.2\%\uparrow}}$ \\
\hline\hline
\end{tabular}
\vskip -0.08in
\caption{ Quantitative comparison for image editing on the PIE benchmark. The methods are compared using P2P attention control~\cite{hertz2022prompt}.  }  \label{tab:p2p}

\end{table*}

\begin{figure}[t]
	
	\begin{center}
		\centering
		{\includegraphics[width=1\columnwidth]{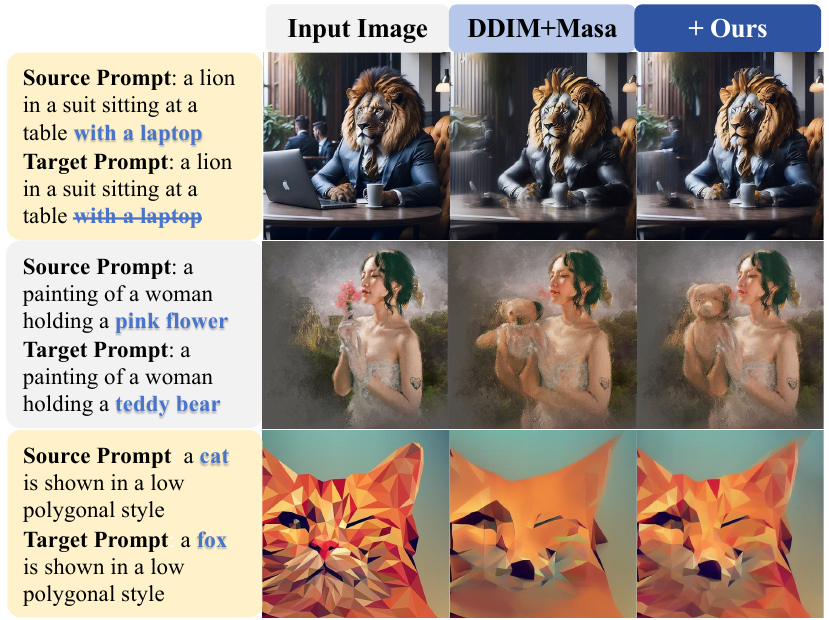}}
		\vskip -0.1in
		\caption{Examples of editing some images using DDIM+Masa.}
		\label{fig:masa_DI}
	\end{center}
	\vskip -0.2in
\end{figure}

\section{Experiments}
\label{sec:experiments}

\begin{figure*}[!ht]
\centering
\begin{minipage}{1\columnwidth}
    \centering
    \includegraphics[width=\columnwidth]{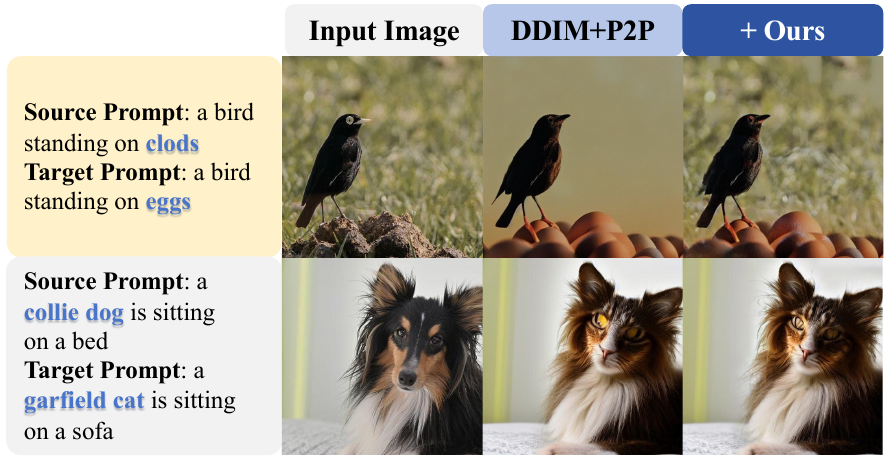}
    \caption{Examples of editing some images using DDIM+P2P on the PIE benchmark.}
    \label{fig:cmp_editing_1}
\end{minipage}
% \hfill
\hspace{0.02\columnwidth}
\begin{minipage}{1\columnwidth}
    \centering
    \includegraphics[width=\columnwidth]{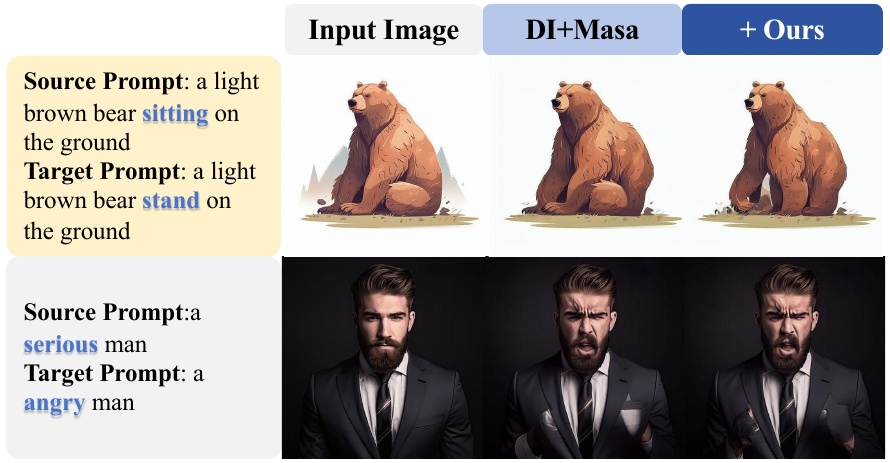}
    \caption{Examples of editing some images using DI+Masa on the PIE benchmark.}
    \label{fig:cmp_editing_2}
\end{minipage}
\vskip -0.18in
\end{figure*}

\begin{figure*}[t]
\centering
\includegraphics[width=2\columnwidth]{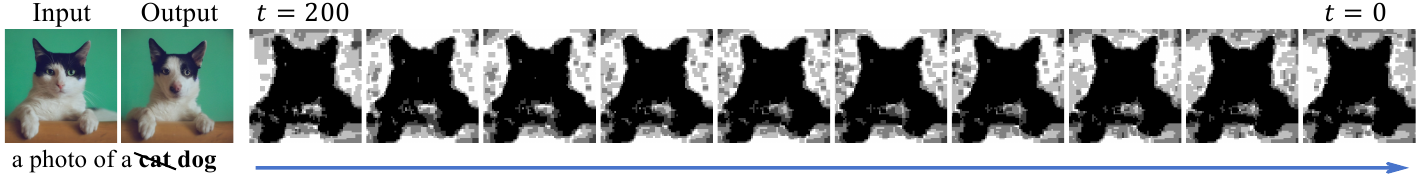} 
\vskip -0.08in
\caption{The adaptive masks generated by our methods.}
\vskip -0.1in
\label{fig:mask}
\end{figure*}

\begin{table}[t]

	\begin{center}
		\resizebox{0.48\textwidth}{!}{
			\begin{tabular}{lcccc}
				\toprule
				Method & $\text{LPIPS}_\text{(BG)}\downarrow$ & $\text{LPIPS}_\text{(FG)}\downarrow$ & $\text{CLIP}_\text{(Image)}\uparrow$ & $\text{CLIP}_\text{(Text)}\uparrow$ \\
				\midrule
				SDEdit (0.4) \cite{meng2021sdedit} & 0.35 & 0.62 & 80.56 & 27.73 \\
				SDEdit (0.6) \cite{meng2021sdedit} & 0.42 & 0.66 & 77.68 & 27.98 \\
				Blended \cite{avrahami2022blended_latent} & 0.11 & 0.77 & 73.25 & 25.19 \\
				Paint \cite{yang2022paint} & 0.13 & 0.73 & 80.26 & 25.92 \\ 
				DIB \cite{zhang2020deep} & 0.11 & 0.63 & 77.57 & 26.84 \\
    				TF-ICON~\cite{DBLP:journals/corr/abs-2307-12493} & {0.10} & {0.60} & {82.86} & {28.11}\\ \hline

        TF-ICON*~\cite{DBLP:journals/corr/abs-2307-12493} & 0.09 &0.51 &80.78 &31.33\\
				\rowcolor[gray]{0.9} + UAM* & \textbf{0.07} & \textbf{0.50} & \textbf{81.10} & \textbf{31.70}\\\hline\hline
				
		\end{tabular}}
	\end{center}
\vskip -0.18in
 	\caption{Quantitative comparison of image composition on {TF-ICON} benchmark~\cite{DBLP:journals/corr/abs-2307-12493}. Additional experimental results and details are in~\cite{DBLP:journals/corr/abs-2307-12493}. Methods marked with `*' indicate results on A800.
  }
	\vskip -0.2in
	\label{tab:quan}
\end{table}

\begin{table*}[t]
\small
    \centering

\begin{tabular}{cl|c|cccc|cc}
\toprule
  &  \multirow{2}{*}{Settings}    & Structure $\downarrow$ & \multicolumn{4}{c}{Background Preservation}& \multicolumn{2}{c}{CLIP Score} \\
 &   & Distance{$\times10^{3}$} & PSNR$\uparrow$ & LPIPS{$\times10^{3}$}$\downarrow$& MSE{$\times10^{4}$}$\downarrow$& SSIM{$\times10^{2}$}$\uparrow$& Whole$\uparrow$  & Edited$\uparrow$ \\\midrule
\multirow{5}{*}{(a)} &$quantile=0.7$ & 21.92 &  23.72 &  80.51 &  66.52 &  82.47 &  24.15 &  20.77 \\
&$quantile=0.6$ &23.35 &  23.30  &  86.37 &  71.69 &  81.78  & 24.27 &  21.15  \\
&$quantile=0.5$ &  24.80 &   22.96  & 91.56 &  76.17&   81.19 &  24.29  & 21.21\\ 
&$quantile=0.4 $&  26.00  &   22.66  & 96.44 &  80.19  & 80.69  & 24.33   &21.24\\ 
&$quantile=0.3$& 26.92  &  22.43  & 100.62 & 83.29 &  80.30   & 24.31  & 21.24\\ \hline

\multirow{3}{*}{(b)} & $T_{mask}=0$ & 28.38 & 22.17 & 106.62 & 86.97 & 79.67&  23.96& {21.16} \\

& $T_{mask}=200$ & 24.80   &  22.96  & 91.56  & 76.17  & 81.19  & 24.29 &  21.21 \\

& $T_{mask}=400$ & 24.70   &  22.96  & 91.85  & 76.03  & 81.11 &  24.28   &21.18\\ 

\bottomrule
\end{tabular} 
\vskip -0.08in
\caption{(a) Ablation study on the influence of $\lambda$ in the editing process using DDIM + Masa with our method when $T_{mask}=200$. (b) Ablation study on the influence of $T_{mask}$ when $quantile=0.5$. }  \label{tab:T_inv}
\vskip -0.1in
\end{table*}

In our experiment, for the image composition task, we follow the experimental setting and composition process of~\cite{DBLP:journals/corr/abs-2307-12493}, using Stable Diffusion v2.1~\cite{rombach2022high} and the 20-step DPM solver sampling method~\cite{Lu2022DPMSolverAF}. We use Uniform Attention Maps (UAM) combined with token values from the target prompts in both the inversion and composition processes. For the image editing task, we follow the setup of~\cite{DBLP:journals/corr/abs-2310-01506}, using the DDIM solver sampling method~\cite{song2021denoising} with 50 steps. The experiments are conducted on a single setup with an A800 GPU, where our method efficiently uses up to 13.7 GB of GPU memory. Additionally, we set the threshold $\lambda$ at the 50\% quantile of the $\mathit{diff}_t$ and $T_{mask}$ to 200, using UAM combined with token values from the null prompts.

\subsection{Experimental Setup}
\noindent{\textbf{Data Set.}} To conduct an objective evaluation of the effectiveness of our method for image editing, we conduct experiments using PIE benchmark~\cite{DBLP:journals/corr/abs-2310-01506}, which has 700 images and a diverse set of complex image editing tasks, including object addition or removal, color changes, and so on. For image composition task, we use the TF-ICON bench mark~\cite{DBLP:journals/corr/abs-2307-12493}. In addition, CelebA-HQ dataset~\cite{karras2017progressive} and PIE benchmark~\cite{DBLP:journals/corr/abs-2310-01506} are used to verify the reconstruction effect of our UAM.

\noindent{\textbf{Comparison to other methods.}}  For the image editing task, we consider several baselines, including DDIM~\cite{song2021denoising}, Null-Text (NT)~\cite{mokady2022null}, Negative Prompt (NP)~\cite{miyake2023negative}, StyleDiffusion (StyleD)~\cite{li2023stylediffusion} and Direct Inversion (DI)~\cite{DBLP:journals/corr/abs-2310-01506}. Additionally, we consider two editing methods: (1) Prompt-to-Prompt (P2P)~\cite{hertz2022prompt} and (2) MasaCtrl (Masa)~\cite{cao2023masactrl}. For the image composition task, we compared our approach with the current state-of-the-art, TF-ICON~\cite{DBLP:journals/corr/abs-2307-12493}.

\subsection{Image Reconstruction}

In \cref{tab:novelty} and \cref{tab:recon_celeb}, our method demonstrates superior reconstruction capabilities, achieving the best results in comparison to the baselines. This further supports the robustness of our approach in generating high-quality images that faithfully adhere to the input specifications.

\subsection{Image Composition}

\noindent{\textbf{Qualitative Evaluation.}} As shown in Fig.~\ref{fig:cmp_tficon}, our method achieves a superior balance between semantic expression and fidelity when compared to TF-ICON~\cite{DBLP:journals/corr/abs-2307-12493}. The visual comparison highlights that our approach not only maintains higher fidelity to the reference images but also produces more coherent and realistic results across diverse contexts, including natural photographs and artistic styles. For instance, in scenarios requiring complex interactions between foreground and background elements, our method successfully preserves the contextual integrity and stylistic consistency, leading to a more harmonious and visually appealing composition. This indicates that our method is particularly effective in handling the subtleties of image composition, where both the content and style need to be accurately represented.

\noindent{\textbf{Quantitative Analysis.}} In Tab.~\ref{tab:quan}, our method consistently outperforms existing approaches across multiple metrics, confirming its effectiveness in image composition tasks. Specifically, our approach achieves the lowest LPIPS scores~\cite{DBLP:conf/cvpr/ZhangIESW18} for both background (LPIPS\(_{BG}\)) and foreground (LPIPS\(_{FG}\)), which indicates a closer perceptual match to the reference images and, therefore, superior visual quality. Additionally, our method exhibits significant improvements in CLIP scores~\cite{Radford2021LearningTV}, with higher CLIP\(_{Image}\) and CLIP\(_{Text}\) values reflecting better alignment between the generated images and the input descriptions. These enhancements suggest that our approach not only excels in producing visually appealing images but also in ensuring that the generated content is semantically coherent and contextually relevant.

\subsection{Image Editing}
\noindent{\textbf{Qualitative Evaluation.}} As shown in Fig.~\ref{fig:masa_DI}, our method demonstrates a superior balance between semantic expression and image fidelity when applied to both real and generated images, outperforming the DDIM+Masa approach. For instance, in the first row, where a lion in a suit is depicted, DDIM+Masa fails to accurately remove the laptop, leaving artifacts that detract from the overall image quality. In contrast, our method successfully preserves the integrity of the original image while effectively applying the desired edits. Similarly, in the second and third rows, our approach maintains the delicate balance between the new and original elements, ensuring that the edits are both contextually appropriate and visually coherent. These examples illustrate that our method better preserves critical image information and mitigates common mismatches or artifacts seen with DDIM+Masa, leading to more realistic and visually appealing results. More results are shown in~\cref{fig:cmp_editing_1,fig:cmp_editing_2}.

\noindent{\textbf{Quantitative Analysis.}} In Tab.~\ref{tab:p2p}, methods enhanced with our approach exhibit superior performance across a range of metrics compared to their baseline counterparts. Specifically, our methods significantly reduce the Structural Distance~\cite{DBLP:conf/cvpr/TumanyanBBD22}, indicating a closer visual resemblance to the original images and thereby enhancing fidelity. Moreover, our approach yields improvements in Background Preservation metrics, as evidenced by increased PSNR and SSIM~\cite{DBLP:journals/tip/WangBSS04} values and decreased LPIPS and MSE scores. These improvements suggest that our method better maintains the original background's integrity while applying the desired edits. Additionally, the CLIP Score for both the whole image and the edited regions shows notable gains, reflecting a more accurate alignment between the generated content and the text prompts. These enhancements collectively underscore the effectiveness of our method in preserving essential image characteristics while performing precise and contextually appropriate edits, thereby achieving a higher quality of image editing compared to existing methods.

\subsection{Visualization of Generated Mask}
In Fig.~\ref{fig:mask}, we illustrate the masks for the cat as shown in Fig.~\ref{fig:method_pipe}. The masks highlight the areas that need modification, and adaptive selection at different time steps ensures that the modifications are not limited to a specific range, resulting in more realistic images. The masks change with each time step, indicating the areas requiring modifications.

\subsection{Ablation Study}
\label{ablation}
\noindent\textbf{Threshold $\lambda$.} As shown in Tab.~\ref{tab:T_inv}~(a), the edited images result from setting the threshold $\lambda$ to different quantiles of the $\mathit{diff}_t$. With an increase in the quantile, the edited image becomes more similar to the original, potentially compromising the desired semantic change.  Consequently, a quantile of $0.5$ is the chosen setting for subsequent experiments because it offers a balance by sufficiently reflecting the target text while preserving a close resemblance to the original image. 

\noindent\textbf{Mask Steps $T_{mask}$.} As shown in Tab.~\ref{tab:T_inv}~(b), we experiment with \( T_{mask} \) values of $0$, $200$, and $400$ for image editing. Notably, \( T_{mask} = 200 \) emerges as the optimal setting, preserving the original image's details while effectively introducing the intended semantic changes. This balance ensures that key features, such as the bear's texture, remain intact while still reflecting the desired alterations. In contrast, when \( T_{mask} = 0 \), the edited image deviates significantly from the original, underscoring the mask's importance. Therefore, we adopt \( T_{mask} = 200 \) for subsequent experiments.

\section{Conclusion}

In this work, we introduce Uniform Attention Maps to replace traditional cross-attention in DDIM-based image reconstruction and editing. Our approach significantly improves the fidelity of image reconstructions while maintaining robustness across different text prompts. We also develop an adaptive mask-guided editing technique that seamlessly integrates with our reconstruction method, enhancing the consistency and accuracy of edits. Experimental results demonstrate that our method outperforms existing approaches in image composition and editing tasks. These findings suggest that Uniform Attention Maps hold strong potential for broader applications in image processing. 

\noindent\textbf{Acknowledgment} This work was supported in part by the National Natural Science Foundation of China No. 62376277, Beijing Outstanding Young Scientist Program NO. BJJWZYJH012019100020098, and Public Computing Cloud, Renmin University of China.

%%%%%%%%% REFERENCES
{\small
\bibliographystyle{ieee_fullname}
\bibliography{egbib}
}

\clearpage
\setcounter{page}{1}
\maketitlesupplementary
% \MakeTitleSupplementary

\section*{A. Adaptive Mask-Guided Image Editing: Algorithm Overview}
The pseudocode for our adaptive mask method is shown in~\cref{algo:mask}. The algorithm takes an input image \( z_0 \), a target prompt \( \mathbf{c}_{tgt} \), and a source prompt \( \mathbf{c}_{src} \). The method starts by inverting the image through auxiliary and source branches and then initializes the target branch from the source branch. 

At each timestep \( t \), we compute noise predictions and update the latent variables in the auxiliary, source, and target branches. It generates an adaptive mask \( M \) by comparing the clean images $\hat{z}_{0}$ from the target and source branches and applies a dilation operation to ensure robustness. The mask \( M \) is then used to blend the predictions from the auxiliary and target branches, preserving key details of the original image while applying the edits.

The process repeats until the final image \( z^{tgt}_0 \) is returned, incorporating the original information and the desired modifications.

\begin{algorithm}[t]
\caption{Edit images with adaptive mask}
\begin{algorithmic}[1]
\STATE {\bfseries Input:} Given original image $z_0$, target prompt $\mathbf{c}_{tgt}$, source prompt $\mathbf{c}_{src}$, denoising model $\boldsymbol{\epsilon}_{\theta}$, uniform cross-attention maps $\mathcal{C}$, null prompt $\mathbf{c}_{\varnothing}$, a dilation operation $dilate(\cdot)$.

\STATE ${z}^u_T \gets \text{Invert}(z_0, \mathcal{C},\mathbf{c}_{\varnothing})$ 
\STATE ${z}^{src}_T \gets \text{Invert}(z_0, \mathbf{c}_{src})$ 
\STATE  ${z}^{tgt}_T \gets{z}^{src}_T$
\FOR{$t = T$ to $1$}
\STATE $\# \ Auxiliary \ Branch $
    \STATE $\epsilon_{u}\gets \boldsymbol{\epsilon}_{\theta}(z^{u}_t, \mathcal{C},\mathbf{c}_{\varnothing})$
    \STATE $\hat{z}^{u}_{0,t} \gets \frac{1}{\sqrt{\alpha_t}}z^{u}_t - \frac{1 - \alpha_t}{\sqrt{\alpha_t}}\epsilon_{u}$
 
\STATE $\# \ Source \ Branch $

        \STATE $\epsilon_{src} \gets \boldsymbol{\epsilon}_{\theta}(z^{src}_t, \mathbf{c}_{src})$
    \STATE $\hat{z}^{src}_{0,t} \gets \frac{1}{\sqrt{\alpha_t}}z^{src}_t - \frac{1 - \alpha_t}{\sqrt{\alpha_t}}\epsilon_{src}$

\STATE $\# \ Target \ Branch $

        \STATE $\epsilon_{tgt} \gets \boldsymbol{\epsilon}_{\theta}({z}^{tgt}_t, \mathbf{c}_{tgt})$

    \STATE $\hat{z}^{tgt}_{0,t} \gets \frac{1}{\sqrt{\alpha_t}}{z}^{tgt}_t - \frac{1 - \alpha_t}{\sqrt{\alpha_t}}\epsilon_{tgt}$

    \STATE $M \gets dilate(|\hat{z}^{tgt}_{0,t} - \hat{z}^{src}_{0,t}| \leq \lambda)$

    \IF{$t < T_{mask}$}

        \STATE $\hat{z}^{tgt}_{0,t} \gets  M \odot \hat{z}^{u}_{0,t} + (1-M) \odot \hat{z}^{tgt}_{0,t} $

    \ENDIF

        \STATE ${z}^{tgt}_{t-1} \gets \sqrt{\alpha_{t-1}}\hat{z}^{tgt}_{0,t} + \sqrt{1 - \alpha_{t-1}}\epsilon_{tgt}$
        \STATE${z}^{src}_{t-1} \gets \sqrt{\alpha_{t-1}}\hat{z}^{src}_{0,t} + \sqrt{1 - \alpha_{t-1}}\epsilon_{src}$
         \STATE $z^{u}_{t-1} \gets \sqrt{\alpha_{t-1}}\hat{z}^{u}_{0,t} + \sqrt{1 - \alpha_{t-1}}\epsilon_{u}$
    
\ENDFOR
\RETURN ${z}^{tgt}_0$
\end{algorithmic}
\label{algo:mask}
\end{algorithm}

\begin{figure}
\centering
\includegraphics[width=1.\columnwidth]{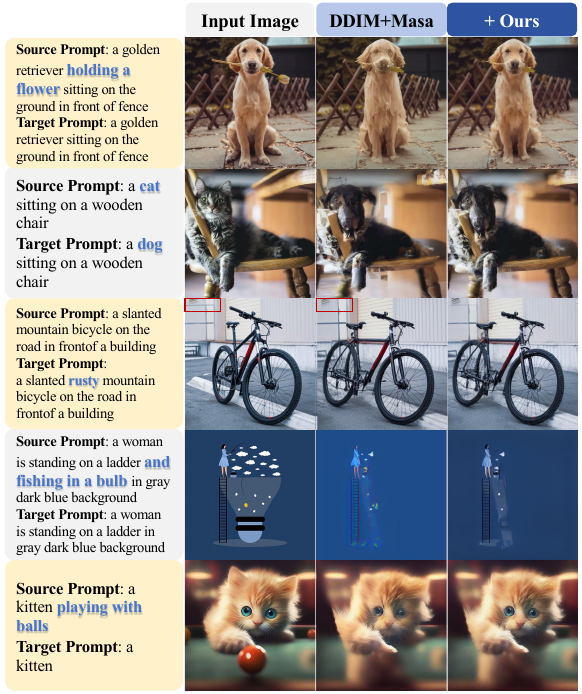} 

\caption{More examples of image editing on the PIE benchmark. Examples of image editing on the PIE benchmark, comparing the DDIM+Masa method with our image editing method. }
\label{fig:sub}
\end{figure}

\section*{B. More Examples of Image Reconstruction}

\begin{figure*}[!ht]
\centering
\includegraphics[width=2.\columnwidth]{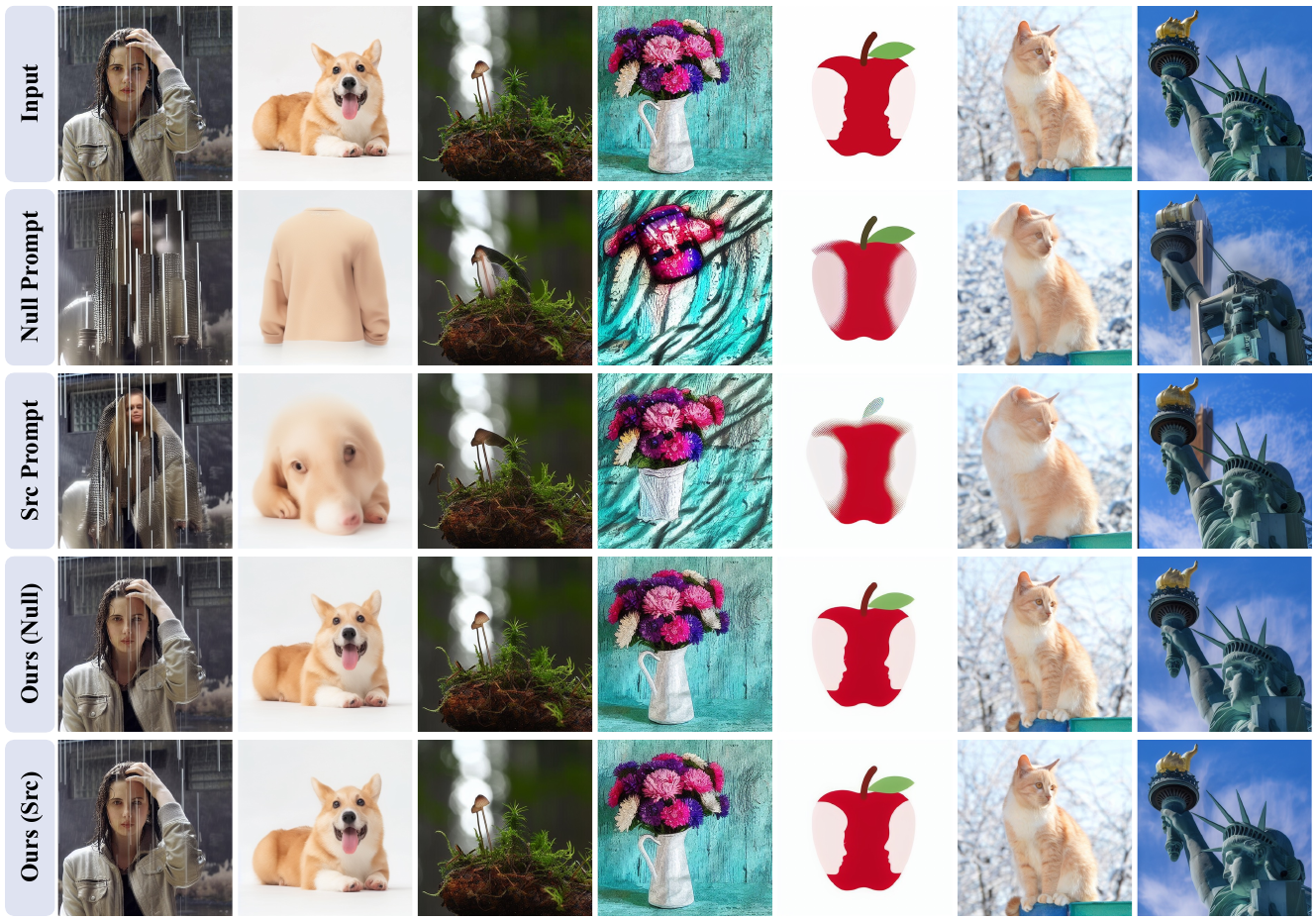} 

\caption{Examples of image reconstruction on the PIE benchmark. The first row shows the input images. The second and third rows display the results using a null prompt (an empty string) and a source prompt from the benchmark, respectively. The fourth and fifth rows show the results from our method with different value tokens, demonstrating superior reconstruction quality and better alignment with the original input images.}
\label{fig:rec_0}
\end{figure*}

\begin{figure*}[!ht]
\centering
\includegraphics[width=2.\columnwidth]{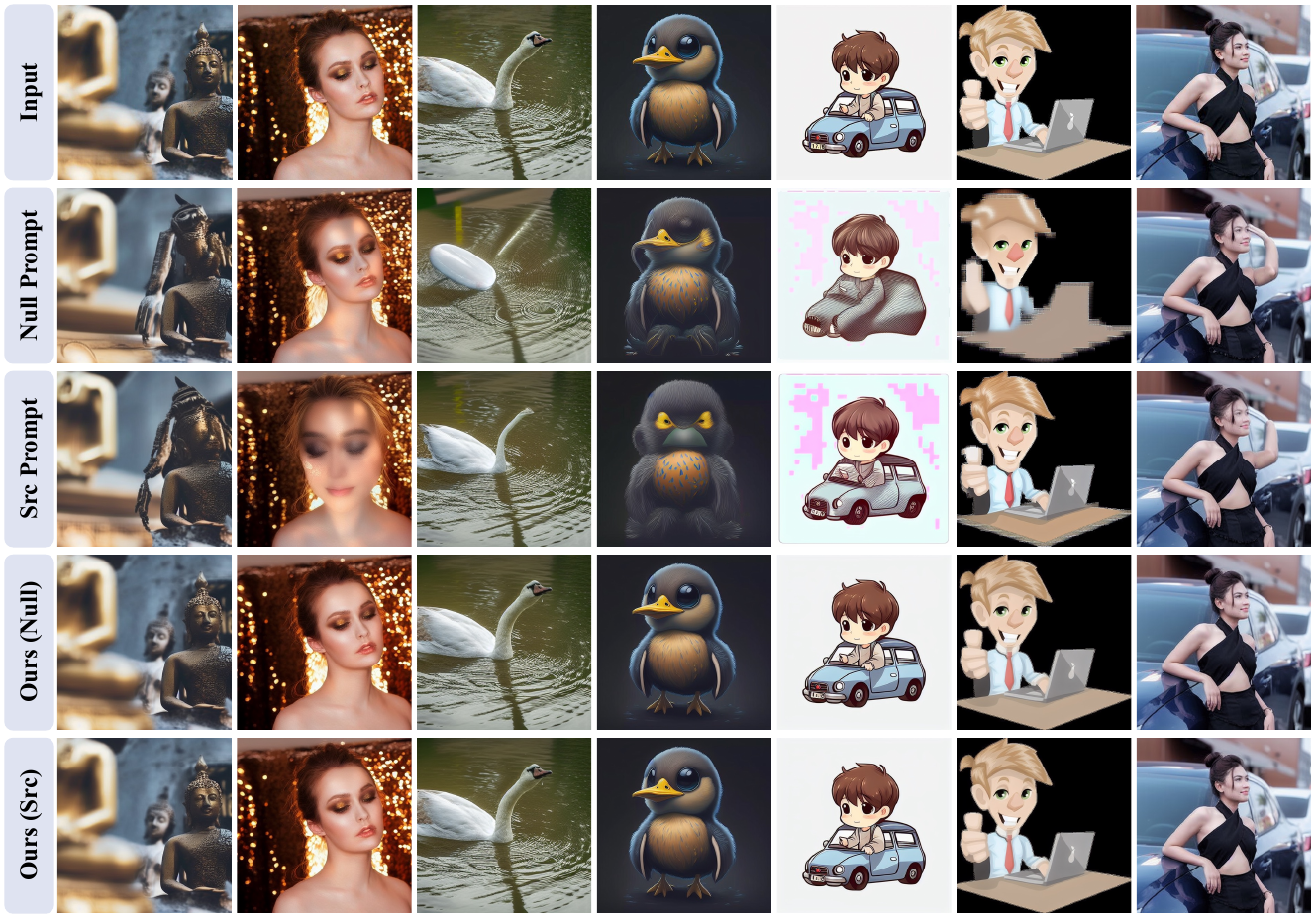} 

\caption{More examples of image reconstruction on the PIE benchmark. }
\label{fig:rec_1}
\end{figure*}

\begin{figure*}[!ht]
\centering
\includegraphics[width=2.\columnwidth]{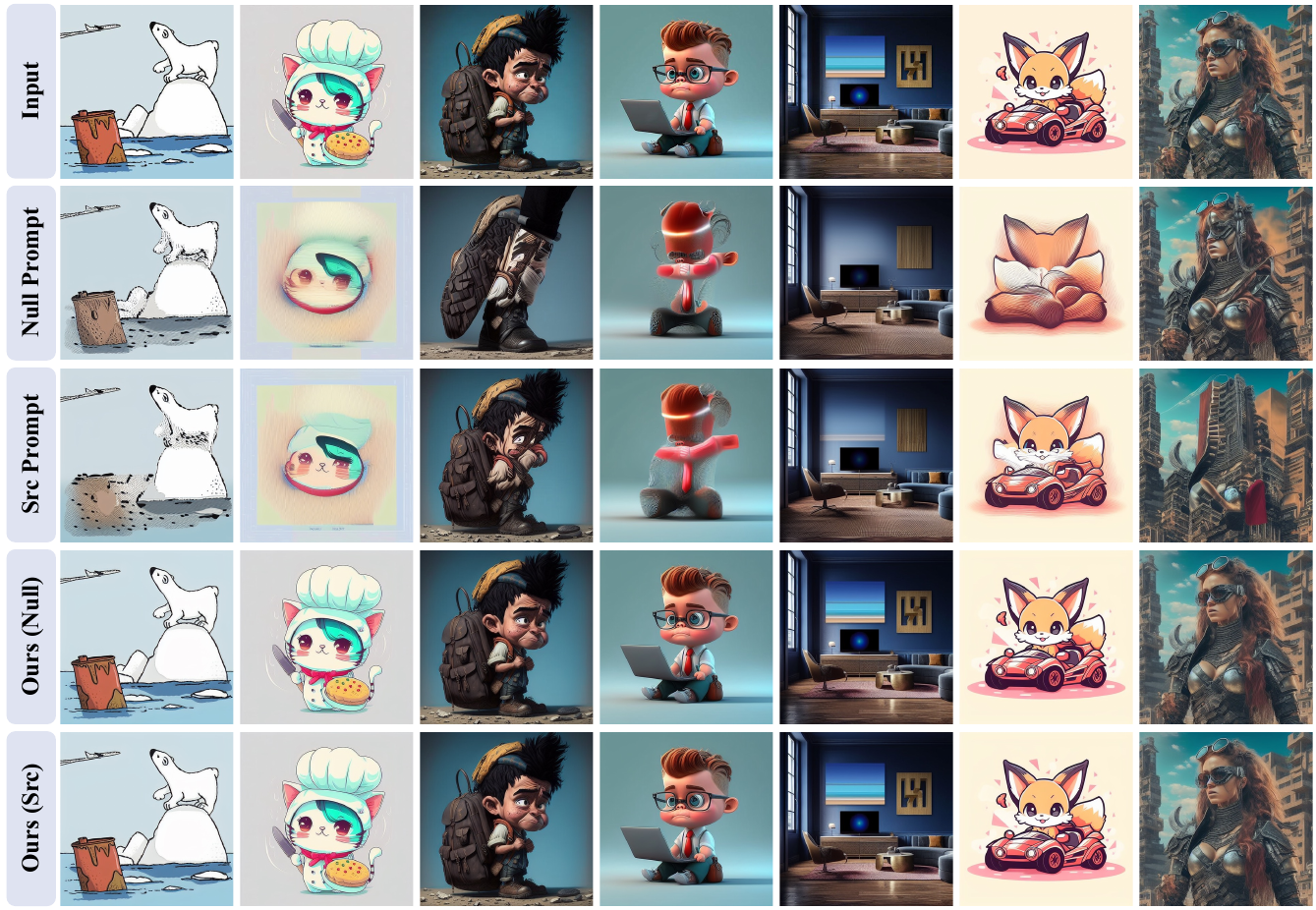} 

\caption{More examples of image reconstruction on the PIE benchmark. }
\label{fig:rec_2}
\end{figure*}

\begin{figure*}[!ht]
\centering
\includegraphics[width=2.\columnwidth]{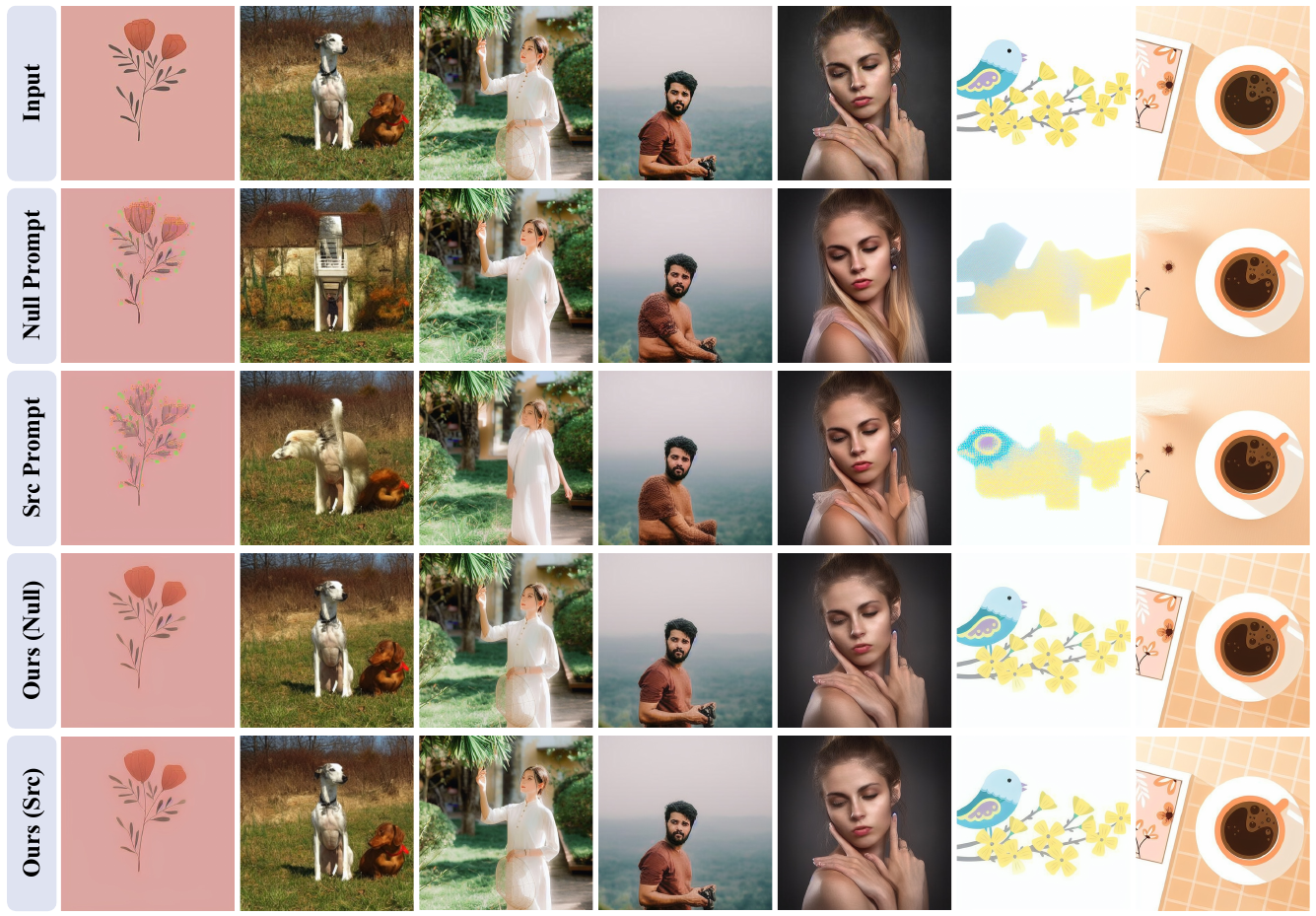} 

\caption{More examples of image reconstruction on the PIE benchmark. }
\label{fig:rec_3}
\end{figure*}

\cref{fig:rec_0,fig:rec_1,fig:rec_2,fig:rec_3}, provide additional examples of image reconstruction using DDIM inversion with $20$ timesteps on the PIE benchmark, showcasing the performance of our method in comparison to null prompts and source prompts. In~\cref{fig:rec_0,fig:rec_1,fig:rec_2,fig:rec_3}, we observe the reconstruction of various images. The results using the null prompt often produce blurred or incorrect outputs, while the source prompt reconstructions are better but still show visible artifacts. By leveraging uniform attention maps, our method demonstrates significant improvements, yielding clearer and more accurate reconstructions that align closely with the original input images, preserving important details such as texture and shape. These examples confirm the robustness of our approach across different image types, showing that our method consistently outperforms the baseline approaches in generating high-quality reconstructions that faithfully resemble the input images.

\section*{C. More Examples of Image Editing}
\cref{fig:sub} showcases the effectiveness of our image editing method compared to the DDIM+Masa baseline. Our method consistently produces more accurate, detailed, and visually coherent edits across various scenarios, such as transforming animals, modifying complex objects, and retaining structural fidelity in abstract compositions, outperforming the baseline in terms of both precision and consistency.

\section*{D.  More Experimental Details}

\noindent\textbf{Visualize Experiment Details.} We conduct experiments in Fig.~\ref{fig:point} and Fig.~\ref{fig:visual_ddim_inv_fail_fruit} using Stable Diffusion v1.4 with DDIM inversion and reconstruction under 20 inference steps. At each timestep, the cross-attention term \( A^{(l)} \) is extracted from U-Net layers with an output dimension of \( 64 \times 64 \). The clean predicted image \( \hat{z}_{0,t} \) is also generated at each timestep $t$ to evaluate the reconstruction fidelity. 

In Fig.~\ref{fig:point}, the Mean Squared Error of the cross-attention term is computed at the pixel level as the discrepancy between \( A^{(l)}_{\text{inv}} \) and \( A^{(l)}_{\text{rec}} \), with the results averaged across all pixels. Similarly, the reconstruction error is calculated as the pixel-level MSE between the predicted clean images \( \hat{z}_{0,\text{inv}} \) and \( \hat{z}_{0,\text{rec}} \). These two MSE metrics are aggregated across all timesteps for each image. The scatter plot in Fig.~\ref{fig:point} illustrates a strong positive correlation between the cross-attention discrepancies and the reconstruction errors, demonstrating that misalignment in the cross-attention mechanism is a significant contributor to the errors in the final reconstructed images.

In Fig.~\ref{fig:visual_ddim_inv_fail_fruit}, the extracted cross-attention terms \( A^{(l)} \) are visualized as heatmaps to show their temporal evolution across the inversion and reconstruction processes. Fig.~\ref{fig:visual_ddim_inv_fail_fruit}~(a) highlights the discrepancies in the cross-attention maps under source prompts, null prompts, and our proposed method. The heatmaps for the source and null conditions reveal significant misalignments between the inversion and reconstruction phases, emphasized by the black-boxed regions. In contrast, our method ensures consistent cross-attention alignment throughout the process.  Furthermore, Fig.~\ref{fig:visual_ddim_inv_fail_fruit}~(b) presents the corresponding clean predicted images  $\hat{z}_{0,t}$  at various timesteps, showing that the proposed method maintains high-quality reconstructions, while the source and null prompts result in noticeable distortions. 

\noindent\textbf{Experimental Metrics.} The primary goal of semantic image editing is to accurately modify specific objects or scenes in an image as described in the target text. This process ensures that only the intended part of the image is altered while retaining unmodified parts as much as possible. To assess the effectiveness of our methods, we utilize metrics from prior work~\cite{DBLP:journals/corr/abs-2310-01506}. We report the following metrics: (1) Peak Signal-to-Noise Ratio (PSNR) and Mean Squared Error (MSE): These metrics evaluate the faithfulness of the generated images by comparing them to the input images. (2) LPIPS~\cite{DBLP:conf/cvpr/ZhangIESW18}: LPIPS is a deep learning-based metric that assesses perceptual similarity between images, aligning more closely with human perception than traditional metrics. (3) SSIM~\cite{DBLP:journals/tip/WangBSS04}: SSIM measures the similarity between the two images, focusing on changes in structural information, luminance, and contrast. (4) CLIP Score~\cite{Radford2021LearningTV}: We employ a combination of CLIP image and text models to calculate the similarity between generated images and corresponding texts, measuring the alignment between the generated image and the target text. We report CLIP Score for both the entire image (Whole) and within the editing mask (Edited), where regions outside the mask are blacked out. (5) Structural Distance~\cite{DBLP:conf/cvpr/TumanyanBBD22}: This metric assesses structural changes in images.

\end{document}